\documentclass[twoside,11pt]{article}

\usepackage{blindtext}

\usepackage{silence}
\usepackage{subcaption}
\usepackage{longtable}

\usepackage[abbrvbib, preprint]{jmlr2e}

\usepackage{lastpage}
\jmlrheading{?}{2025}{1-\pageref{LastPage}}{5/25}{?}{?}{Connor Cooper, Geoffrey I. Webb, Daniel F. Schmidt}

\ShortHeadings{Efficient Parameter Estimation for BNCs}{Cooper, Webb, and Schmidt}
\firstpageno{1}

\usepackage{amsmath}
\usepackage{bm}
\usepackage{tikz}
\usepackage{multirow}
\usepackage{parskip}
\usepackage[T1]{fontenc}
\usepackage{booktabs}
\usepackage[misc]{ifsym}
\usepackage{booktabs}   
\usepackage{float}

\begin{document}

\title{Efficient Parameter Estimation for Bayesian Network Classifiers using Hierarchical Linear Smoothing}

\author{\name Connor Cooper \email connor.cooper1@monash.edu \\
       \name Geoffrey I. Webb \email geoff.webb@monash.edu \\
       \name Daniel F. Schmidt \email daniel.schmidt@monash.edu \\
       \addr Faculty of Information Technology\\
       Monash University \\
       Melbourne, Australia
       }

\editor{}

\maketitle 

\begin{abstract}%
Bayesian network classifiers (BNCs) possess a number of properties desirable for a modern classifier: They are easily interpretable, highly scalable, and offer adaptable complexity. However, traditional methods for learning BNCs have historically underperformed when compared to leading classification methods such as random forests.
Recent parameter smoothing techniques using hierarchical Dirichlet processes (HDPs) have enabled BNCs to achieve performance competitive with random forests on categorical data, but these techniques are relatively inflexible, and require a complicated, specialized sampling process.
In this paper, we introduce a novel method for parameter estimation that uses a log-linear regression to approximate the behaviour of HDPs. As a linear model, our method is remarkably flexible and simple to interpret, and can leverage the vast literature on learning linear models.
Our experiments show that our method can outperform HDP smoothing while being orders of magnitude faster, remaining competitive with random forests on categorical data.

\end{abstract}

\section{Introduction}
As the availability of large and complex datasets grows,
Bayesian network classifiers (BNCs) (see Friedman et al., 1997, for example) are a class of models that offer a number of properties desirable for modern classifiers:
\begin{itemize}
  \item As Bayesian networks, the evaluation and interpretation of predictions can be formally and precisely derived
  \item They provide a convenient mechanism for controlling the complexity of the model, and this complexity can be tuned to the data
  \item They can be learned without holding all the data in memory
  \item The parameters for each node are conditional probability tables (CPTs), which can each be learned independently
  \item They can scale linearly with the number of predictors, assuming a bounded number of parents for each node.
\end{itemize}
However, BNCs are typically outperformed by leading classification methods such as random forests.
Parameters for each CPT are usually estimated individually, using simple maximum likelihood estimates or additive smoothing, without consideration to how neighbouring cells in the CPTs might be correlated.
In situations where some parameters in the CPTs have small numbers of samples, these estimates often fail to deliver accurate results. Due to the combinatorial number of parameters in a CPT, this can happen very often, even for an overall large amount of data.
In response there has been some attention paid to methods that estimate the parameters of the CPTs jointly, by sharing data between parameters.
Notably, \citet{petitjean_accurate_2018} and \citet{azzimonti_hierarchical_2019} have found success using hierarchical Dirichlet processes (HDPs) to smooth these parameter estimates for categorical data.
These methods treat each CPT as a tree that branches on the values of each parent of the child node, and assign shared Dirichlet priors in a hierarchy according to the CPT's tree.
This has been shown to outperform additive smoothing, and can make BNCs competitive with leading methods such as random forests, even for relatively simple structures.
However, HDP smoothing is limited in its application due to requiring a complicated and relatively inefficient sampling process that is inflexible to changes or extensions to the sharing behaviour.

In this paper, we present an alternative method for joint learning of the parameters of Bayesian network classifiers on categorical data. 
Our method reformulates HDP smoothing as a log-linear model, which results in a model that is both simple and computationally efficient.
By leveraging existing theory for linear models, we can avoid the shortcomings of HDP smoothing while introducing many of the advantages of linear models.
We do this by taking the hierarchical tree structure of \citet{petitjean_accurate_2018} (see Figure 1a), and representing it as a design matrix in a multinomial logistic regression, with a predictor for each node in the tree, where the targets are the parameters we wish to estimate.
Parameter estimates can then be obtained using any method for learning linear models.
We refer to this model as hierarchical linear smoothing (HLS).
This approach also shares some similarities with \citet{rijmen_bayesian_2008}.

The key contribution of this paper is our log-linear model for joint estimation of conditional probability tables. Specifically, in Section 3 we present our log-linear formulation of HDP smoothing, and in Section 4 perform extensive experiments demonstrating its computational efficiency and alignment to HDP smoothing. We also demonstrate our formulation's competitiveness with random forests.
Code for using HLS, as well as running our experiments and loading datasets, is available at \url{https://github.com/coopco/HLS}.

\section{Bayesian networks}
The following framework can be found in texts such as \citet{friedman_bayesian_1997}.
A Bayesian network $\mathcal{B}=(G, \Theta)$ is defined by its structure $G$ and set of parameters $\Theta$. The structure is a directed acyclic graph, where the nodes are random variables $\{X_i\}_{1\leq i\leq p}$, and edges represent dependencies between variables. 
For each node $X_i$ with parents denoted by $\Pi_{X_i}$, $\Theta$ contains a conditional probability table (CPT) representing the distribution of $X_i$ given $\Pi_{X_i}$.
For each possible configuration $x_i$ of $X_i$ and $\Pi_{x_i}$ of $\Pi_{X_i}$, the CPT contains the parameter $\theta_{x_i\,|\,\Pi_{x_i}} := P_{\mathcal{B}}(x_i\,|\,\Pi_{x_i})$.
The Bayesian network than estimates the full joint probability distribution over $\mathbf{x}$ as \begin{equation*} 
  P_B(\mathbf{x}) = \prod_{i=1}^{p} P_B(x_i \,|\, \Pi_{x_i}) = \prod_{i=1}^{p} \theta_{x_i \,|\, \Pi_{x_i}}.
\end{equation*}
A Bayesian network classifier extends the Bayesian network to model the distribution of a target class variable $Y$ conditioned on the variables $\{X_i\}_{1\leq i\leq p}$, so we add the additional node $Y$ to the network and write \begin{equation*} 
  P_B(y\,|\,\mathbf{x}) = \frac{\theta_{y \,|\, \Pi_y} \prod_{i=1}^{p} \theta_{x_i \,|\, \Pi_{x_i}}} {\sum_{y'\in \mathcal{Y}} \theta_{y' \,|\, \Pi_{y'}} \prod_{i=1}^{p} \theta_{x_i \,|\, \Pi_{x_i}}},
\end{equation*} where $\mathcal{Y}$ is the sample space of $Y$ and $y\in \mathcal{Y}$.
Training a Bayesian network classifier thus requires two steps:
\begin{enumerate}
  \item Learning the structure of the network $G$, i.e. the dependencies between variables, and
  \item Learning the set of parameters $\Theta$, i.e. the conditional probability tables for each node.
\end{enumerate}

\subsection{Structure learning}
The complexity of a Bayesian network classifier can be thought of as being controlled by the number of edges in the network. As the number of parameters in a node's CPT scales exponentially with the number of parents, successful structure learning methods must make careful consideration of what edges to include in the network. For recent BNCs, this is usually done discriminatively.
There is a large variety of structure learning methods in the literature,
but to remain focused on parameter learning we restrict our attention to two classes of structures we use in our experiments.

\subsubsection{Tree-augmented Bayes}

Tree-augmented Bayes (TAN) \citep{friedman_bayesian_1997} uses the pair-wise mutual information between attributes as edge weights to find a maximum spanning tree over the non-class nodes. The class node is added as a parent to every node.

\subsubsection{K-dependence Bayes}
K-dependence Bayes \citep{sahami_learning_1996} distills the complexity trade-off into a single
hyperparameter $K$, which determines the maximum number of parents for each
node.
The parents chosen for each node are the class, along with the $K$
attributes with the highest mutual information conditioned on the class, among
those with higher mutual information with the class.
We later refer to K-depence Bayes with hyperparameter $K$ as kDB-$K$.

\subsection{Parameter learning} \label{parameterlearning}

Typically, each parameter in each CPT is estimated individually.
The simplest choice of method uses the maximum likelihood estimates, which are just the sample means from the data:
\begin{equation*}
  \theta_{x_c\,|\,\Pi_{x_c}} = \frac{n_{x_c\,|\,\Pi_{x_c}}}{n_{\Pi_{x_c}}},
\end{equation*}
where $n_{x_c\,|\,\Pi_{x_c}}$ is the observed frequency of $X_c = x_c$ when $\Pi_{X_c} = \Pi_{x_c}$, and $n_{\Pi_{x_c}}$ is the total number of observations where $\Pi_{X_c} = \Pi_{x_c}$.
Smoothing can be applied by assigning a Dirichlet prior to each parameter \begin{equation*} 
  \theta_{x_c\,|\,\Pi_{x_c}} \sim \text{Dir}(m).
\end{equation*}
This is equivalent to adding $m$ pseudo-counts to each estimate \begin{equation*} 
  \theta_{x_c\,|\,\Pi_{x_c}} = \frac{n_{x_c\,|\,\Pi_{x_c}} + m}{n_{\Pi_{x_c}} + |X_c|\cdot m}.
\end{equation*}
Choices of $m$ correspond to common estimates: $m=0$ corresponds to the maximum likelihood estimates, $m=1/2$ corresponds to Jeffery's prior, and $m=1$ corresponds to a uniform prior \citep{wang_horseshoe_2024}.

\begin{figure*}[t]
  \centering
  \begin{subfigure}[b]{0.3\textwidth}
    \centering
    \includegraphics[width=0.9\textwidth]{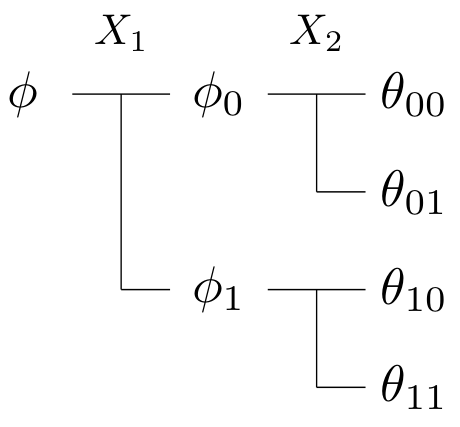}


    %

    \caption{HDP prior structure}
    \label{fig:hdp}
  \end{subfigure}
  \hfill
  \begin{subfigure}[b]{0.3\textwidth}
    \centering
    \includegraphics[width=0.9\textwidth]{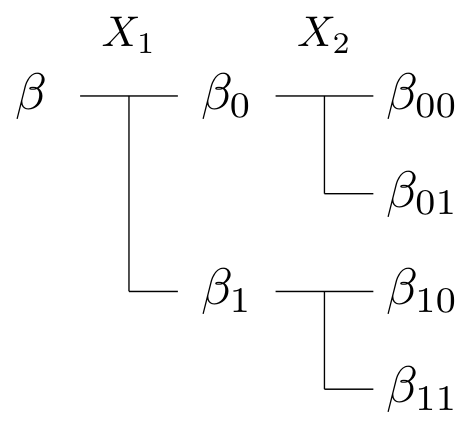}



    \caption{HLS coefficients}
    \label{fig:hls}
  \end{subfigure}
  \hfill
  \begin{subfigure}[b]{0.3\textwidth}
    \centering
    \includegraphics[width=\textwidth]{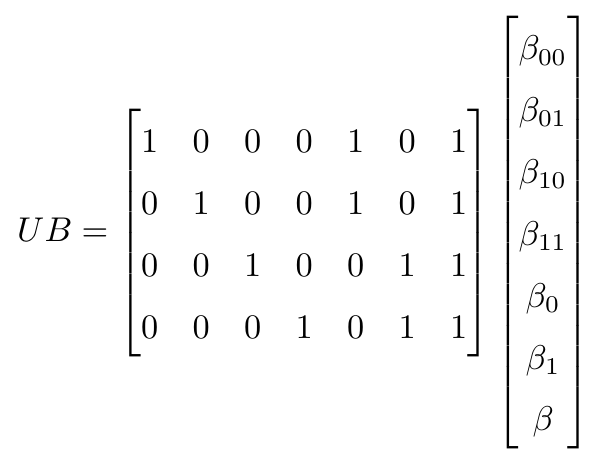}
    \caption{HLS design matrix}
    \label{fig:design}
  \end{subfigure}
  \caption{For a child node $X_c$ with binary-valued parents $X_1$ and $X_2$,
    a tree that branches first on the values of $X_1$ and then on the values of $X_2$, with the corresponding:
    (a) parameters for HDP, (b) coefficients for HLS, and
    (c) design matrix for HLS.}
  \label{fig:three graphs}
\end{figure*}

\subsubsection{Joint learning of parameters}
There is an expectation of some degree of smoothness to the cells of a CPT. 
For example, you might expect the parameter for, say, $P(X_c \,|\, X_1=0, X_2=0, X_3=0)$ 
to be more similar to $P(X_c \,|\, X_1=0, X_2=0, X_3=1)$ than it is to, say, $P(X_c \,|\, X_1=1, X_2=1, X_3=1)$.
This suggests that there is potential to estimating the parameters of a CPT jointly by sharing data between parameters.
One can then think of a hierarchy where, given a node and an ordering of its parents, the CPT is represented as a tree that branches on the possible values of the first parent, then on the possible values of the second parent, and so on for all parents of the node. The distance between leaves in this tree then models an {\em a priori} expectation of similarity between the leaf distributions.
This is the core idea behind the approach from \citet{petitjean_accurate_2018}, where this hierarchy is modelled using hierarchical Dirichlet process (HDP) priors shared between the parameters.
This same HDP smoothing idea has also been extended to ensembles of networks \citep{zhang_bayesian_2020} and probability estimation trees \citep{zhang_hierarchical_2020}.

In the above example, one would expect the parameters for $P(X_c \,|\, X_1=0, X_2=0, X_3=0)$ and $P(X_c \,|\, X_1=0, X_2=0, X_3=1)$ (say, $\theta_1$ and $\theta_2$ for short) to both be similar to $P(X_c \,|\, X_1=0, X_2=0)$. In this case, you introduce a latent prior parameter $\phi$ for $\theta_1$ and $\theta_2$: 
\begin{align*}
  \theta_{1} \sim \text{Dir}(\phi, \alpha_1) \\
  \theta_{2} \sim \text{Dir}(\phi, \alpha_2),
\end{align*}
for some concentrations $\alpha_1$ and $\alpha_2$.\footnote{Here, we use a two-parameter $\text{Dirichlet}(\phi,\alpha)$, where the usual vector parameter is split into a base probability vector $\phi$ and scalar concentration $\alpha$}
This hierarchy extends upwards for all internal nodes of the tree. Figure 1(a) shows an example hierarchical structure for a node with two parents.
The final hierarchy is specified by:
\begin{align*}
\theta_{X_c\,|\,y,x_1,\dots,x_{p}} &\sim \text{Dir}(\phi_{X_c\,|\,y,x_1,\dots,x_{p-1}},\, \alpha_{y,x_1,\dots,x_p}) \\
\phi_{X_c\,|\,y,x_1,\dots,x_i}   &\sim \text{Dir}(\phi_{X_c\,|\,y,x_1,\dots,x_{i-1}},\, \alpha_{y,x_1,\dots,x_{i}}) \\
\phi_{X_c\,|\, y} &\sim \text{Dir}(\phi_{X_c},\alpha_y) \\
\phi_{X_c}   &\sim \text{Dir}\left(\frac{1}{|X_c|}\mathbf{1},\, \alpha_{0}\right),
\end{align*} 
for each realization $(y, x_1, \dots, x_p)$ of $\Pi_{X_i}$, for $i=1,\dots,p-1$.
The $\alpha$ parameters are often given a gamma prior \citep{zhang_bayesian_2020}.
The result is that each leaf has a parameter vector $\theta_{X_c\,|\,y,x_1,\dots,x_n}$, and an internal node at depth $i$ has a latent prior parameter vector $\phi_{X_c\,|\,y,x_1,\dots,x_i}$. 
The values of the parameters are found via a complex and highly intricate Gibbs sampling approach.

\section{Hierarchical Linear Smoothing}
Our goal is to estimate each of the $p$ conditional probability tables of a Bayesian network.
For each CPT, we must estimate a parameter for each possible configuration of the parents. We do this by constructing a design matrix for each CPT that allows the parameters to be learned through a logistic regression.
Given this design matrix, any method for training a logistic regression can be then used to learn the CPTs.

In this section, we first describe the design matrix that allows the CPT parameters to be expressed as targets in a logistic regression.
Following that, we discuss the procedures we use in our experiments for fitting the logistic regressions.

\subsection{Design matrix}

Each node in the network can be estimated independently, so without loss of generality, we consider a child node $X_c$ with parents $\Pi_{X_c}$ ordered as $(X_1,\dots,X_p)$.
We proceed similarly to HDP smoothing, treating the conditional probability table for child node $X_c$ as a tree that branches on the values of each parent in $\Pi_{X_c}$. Parents in $\Pi_{X_c}$ are ordered based on mutual information with the child conditioned on the class $Y$.
Let $N$ be the number of nodes in the corresponding tree, and $L$ be the number of leaves.
Our goal is to estimate the $L$ parameters corresponding to the leaves of this tree, which is done in a logistic regression framework.

We do this by introducing coefficients $\beta_j$ for each node $j$ in the tree.
A linear predictor is then defined for each leaf node as the sum of the coefficients corresponding to each of its ancestors in the tree:
\begin{equation*}
  \eta_{X_c\,|\, (x_1, \dots, x_p)} = \sum_{i=1}^{p} \beta_{x_1,\dots,x_i},
\end{equation*} for each of the $L$ leaves.
With the coefficients collected in an $N\times |X_c|$ matrix $B$, and the linear predictors collected in an $L\times |X_c|$ matrix $H$, the linear predictors are formed using the $L\times N$ design matrix $U$
\begin{align*}
H &= UB, \\
H = \begin{bmatrix}
    \eta_{1} \\ \vdots \\ \eta_{L}
\end{bmatrix}&,\quad
B = \begin{bmatrix}
    \beta_{1} \\ \vdots \\ \beta_{N}
\end{bmatrix},
\end{align*}
where the coefficients in $B$ are ordered by ascending depth. The order of coefficients within a depth is unimportant, so long as it is consistent with $H$. $U$ can be defined to represent the CPT tree by
\begin{equation*}
  U_{ij} = \begin{cases}
  1 \quad\text{if $i=j$ or $j$ is an ancestor of $i$,}  \\
  0 \quad\text{otherwise.}
  \end{cases}
\end{equation*}
In Figure 1, we give an example tree (b) and design matrix (c) for a node with two binary-valued parents.

The idea underlying this approach is that the coefficients are shrunk towards the values of the coefficients of their ancestors. For example, if the effect of $X_p$ given $X_1,\dots,X_{p-1}$ is very small, than the corresponding $\beta$ coefficients should be shrunken, while the coefficients for, say, $X_{p-1}$ given $X_1,\dots,X_{p-2}$ are promoted.
%
%

Depending on how the regression is framed, the targets will be either multinomial or categorical.
In the categorical case, the rows of the design matrix will have to be repeated for each count.
In practice, we also drop the final column of the design matrix, as it functions similarly to an intercept.

\subsection{Regression} \label{glsh}

%

Given an estimate of the coefficients $B$, and the corresponding linear predictors $H = UB$, the CPT parameters can be computed via
\begin{equation*}
  \theta_{x_c\,|\,\Pi_{x_c}} = \text{softmax}(\eta_{x_c \,|\, \Pi_{x_c}}),
\end{equation*} for all configurations of $x_c$ and $\Pi_{x_c}$.
Given the design matrix $U$, we can learn all the parameters $B$ of the CPT jointly through a standard logistic regression.
For instance, it is common to estimate the coefficients using penalized maximum likelihood, where a penalty term
(such as the ridge penalty $\tau \lvert\lvert \beta \rvert\rvert^2_2$ or the lasso penalty $\tau \lvert\lvert \beta \rvert\rvert_1$)
is added to the likelihood function \citep{van_wieringen_lecture_2023}.
However, there are some important considerations that come with selecting a method:
\begin{itemize}
  \item An implementation using sparse matrices is vital to maintain the efficiency of HLS: the design matrix is contains only $L \times p$ non-zero elements
  \item Similarly, as mentioned above, an implementation that allows duplicate categorical counts to be collected into multinomial counts is also vital for large datasets.
  \item Using a method such as cross-validation (CV) to select the regularization strength does not work well in this setting. The nature of the problem means that we are often working with sparse design matrices, which can result in high variances cross-validation estimates of risk, and subsequently sub-optimal selection of the regularization hyperparameter. This weakness of cross-validation has been discussed in \citet{tew_bayes_2023}.
\end{itemize}

Nevertheless, common logistic ridge regression solvers (such as in Scikit-learn \citep{pedregosa_scikit-learn_2011}, which we use in our experiments) with a fixed regularization parameter $\tau$ provide a strong baseline (see Section \ref{nonbayes}). 

As an alternative to cross-validation, we also experiment with more flexible Bayesian penalized regression procedures. The aim is to determine what improvements can be found from optimising the degree of regularization.
To this end, we choose to use the global-local shrinkage (GLS) hierarchy \citep{polson_local_2011}.
To model our multinomial regression under this framework, the hierarchy (see \citealp{makalic_high-dimensional_2016}) is given by
\begin{align*}
  \theta_i \,|\, \mathbf{u}_i, w_i^2 &\sim N(\mathbf{u}_i^T B, w_i^2) \\
  w_i^2 &\sim \text{PG}(0, 1) \\
	\beta_j | \lambda_j^2, \tau^2, &\sim N(0, \lambda_j^2\tau^2) \\
	\lambda_j &\sim p(\lambda_j) d\lambda_j \\
	\tau &\sim p(\tau) d\tau 
\end{align*} for $1\leq i\leq L$ and $1\leq j\leq N$, where $N(\mu, \sigma^2)$ is the normal distribution with mean $\mu$ and variance $\sigma^2$, and $\mathbf{u}_i$ is row $i$ of the design matrix $U$.
The logit target is modelled through a scale mixture of normals: the scales $w_i \sim \text{PG}(0, 1)$ follow a standard P\'olya-Gamma distribution \citep{polson_bayesian_2013} which results in a logistic likelihood.
The prior distributions $p(\lambda_j)$ and $p(\tau)$ control the behaviour of the shrinkage estimator, and specific choices lead to standard Bayesian shrinkage procedures.
In particular:
\begin{itemize}
  \item $\lambda_j = 1$ and $p(\tau) \sim C^+(0, 1)$ corresponds to the Bayesian ridge, where  $C^+(0, 1)$ is the standard half-Cauchy distribution.
  \item $p(\lambda_j) \sim C^+(0, 1)$ and $p(\tau) \sim C^+(0, 1)$ corresponds to the horseshoe \citep{carvalho_horseshoe_2010}.
\end{itemize}

In this paper, we use the Bayesian ridge under this framework. Results using the Horseshoe are discussed in Appendix A.4.
Following \citet{makalic_high-dimensional_2016}, we sample the coefficients from $p(\beta\,|\,\lambda,\tau)$ and $p(\omega\,|\,\beta)$ using the algorithm from Rue (\citeyear{rue_fast_2001}).
For the shrinkage hyperparameters, we sample from  
$p(\tau\,|\,\beta)$ using the algorithms of \cite{schmidt_bayesian_2020}. To sample from the P\'olya-Gamma distribution, we use a tail-corrected gamma series described in Appendix B. This strength of this sampler is its efficiency for large values of $n$.

\begin{figure*}[t]
     \centering
    \begin{subfigure}[b]{0.55\textwidth}
      \centering
      \begin{tabular}{ccc}
      \hline
      \multicolumn{3}{c}{HLS regularization vs Ridge ($\tau=1$)} \\
      \multirow{2}{*}{Regularization} & \multicolumn{2}{c}{Win-draw-loss} \\
                       & 0/1-loss & Log loss \\ \hline
      Ridge with CV         & 5-2-43 & 9-0-41 \\
      Ridge with improved CV & 18-6-26 & 17-0-33 \\
      Lasso            & 26-3-21 & 22-0-28 \\
      Half-Cauchy           & 19-6-25 & 24-0-26 \\
      Inverse-gamma               & 25-7-18 & 35-0-15 \\ \hline
      \end{tabular}
      \caption{}
      \label{fig:regularization}
    \end{subfigure}
    \begin{subfigure}[b]{0.4\textwidth}
        \centering
        \includegraphics[width=\textwidth]{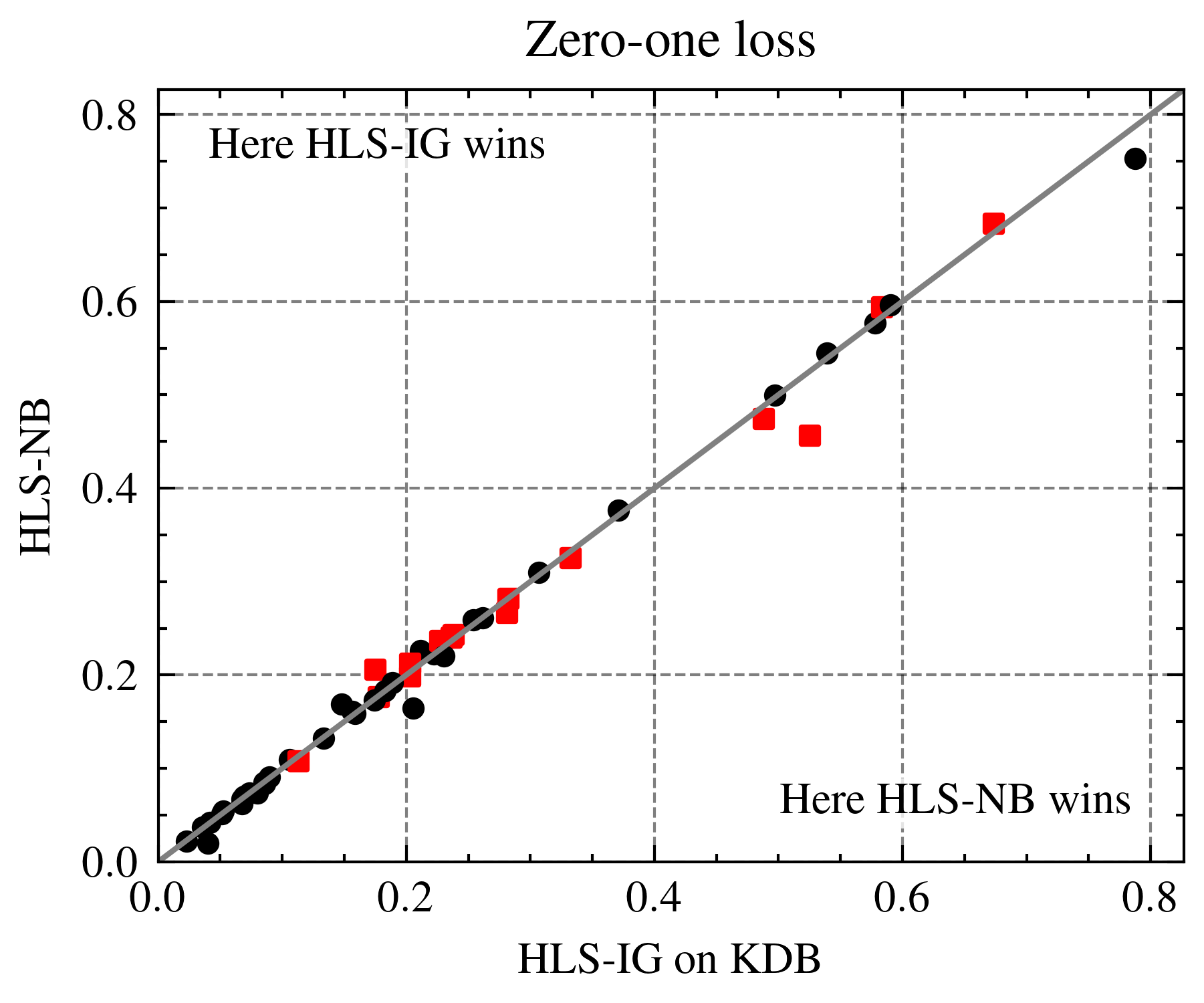}
        \caption{}
        \label{fig:igvnb1}
    \end{subfigure}
    \caption{(a) Win-Draw-Loss records on kDB-3 for various regularization strategies vs ridge with $\tau=1$. (b) Scatter plot for Bayesian inverse-gamma (HLS-IG) vs ridge regression (HLS-NB) on kDB-3, under zero-one loss. Red squares indicate datasets with top-15 CV variance.}
    \label{fig:igvnb}
\end{figure*}


\section{Experiments}

The aim of this section is two-fold: we wish to explore the problem of estimating our regression coefficients,
and to assess how BNCs using our HLS estimates perform.
In Section \ref{setting} we describe the data we use and how models are assessed.
Then, in Section \ref{implementations} we list the implementations and hyperparameters of any models we use.
Section \ref{ablation} concerns the issue of learning estimates of our HLS regression coefficients,
through non-Bayesian methods in Section \ref{nonbayes} and Bayesian methods in Section \ref{bayes}.
in Section \ref{performance} we assess the performance of our HLS estimates: 
Section \ref{bncs} compares HLS to other parameter learning methods for BNCs, and Section \ref{rf} compares our BNCs against random forests.
Finally, in Section \ref{timing} we report the runtime for our models on real and simulated data.

\subsection{Setting} \label{setting}
For our experiments, we use a collection of 50 datasets from the UCI archive \citep{lichman_uci_2013}: 28 datasets with less than 1000 samples, 18 datasets with number of samples between 1000 and 10000, and 4 datasets with more than 10000 samples.
A full list of the datasets is given in Appendix C.
For all methods, numerical attributes are discretized according to the minimum description length principal \citep{fayyad_multi-interval_1993}, using the implementation in \citet{lin_hlin117mdlp-discretization_2017}.
For each method and each dataset, we test using 10-fold cross validation.
When comparing two methods, we use the zero-one classification loss and the log loss, and report their overall Win-Draw-Loss (W-D-L) across all datasets. The variance over cross-validation folds can be very high for some datasets, so we highlight those points in our scatter plots.

\subsection{Implementations} \label{implementations}
For our non-Bayesian regression models, we use the implementations in the Scikit-learn \citep{pedregosa_scikit-learn_2011} library for Python, with sparse matrices for the designs. Cross-validation also uses the Scikit-learn implementation. We also do not include an intercept, as it does not seem to impact performance. Also note that Scikit-learn is inefficient in the sense that it does not use multinomial counts for categorical data (see Section \ref{glsh})
The samplers for the GLS hierarchy (see Section \ref{glsh}) are implemented in Python, using Scipy sparse matrices. Our sampler for the P\'olya-Gamma distribution is implemented in Cython.

For additive smoothing, if a leaf parameter has zero samples, we use the nearest back-off estimate that has at least one sample.
For example, if $n_{x_c\,|\,x_1, x_2}=0$, we consider consider $\hat{p}(x_c\,|\, x_1)$ instead of $\hat{p}(x_c\,|\, x_1, x_2)$.
Not doing this is known to significantly degrade the performance of additive smoothing \citep{petitjean_accurate_2018}.
For HDP smoothing, we use the implementation provided by \cite{petitjean_accurate_2018},
with the default choices of hyperparameters.

For random forests, we use the implementation in the Scikit-learn package \citep{pedregosa_scikit-learn_2011}. We train 100 trees in each forest, with the square root of the number of features considered at each split, using the Gini impurity criterion (see documentation for \citealp{pedregosa_scikit-learn_2011}).
It is important to note that random forests are trained on the same discretized data that our BNCs are trained on; the goal is to compare the methods on categorical data, not to test the effectiveness of the discretization method.

\subsection{HLS experiments} \label{ablation}

In this section we look at several different methods for learning regularized estimates of our HLS regression coefficients.
Results in this section are presented for kDB-3, as it provides larger networks (compared to TAN and smaller kDBs) that should better show the difference in behaviour between methods.

\subsubsection{Non-Bayesian} \label{nonbayes}

We first look at using cross-validation (CV) to select a shrinkage parameter $\tau$ for ridge regression.
From the results in Table \ref{fig:regularization}, we see that the Scikit-learn implementation actually fails to provide reasonable results.
By noting that the best values of $\tau$ tend to be fairly small, we can improve CV by choosing $\tau$ from a small interval $\tau \in [0, 5]$. However, this still performs worse than simpler procedures.
As mentioned in Section \ref{glsh} this could be attributed to the sparsity of the counts: with many parameters having very few samples, the variance of each cross-validation step can be quite high.
Similar behaviour has been discussed in \citet{tew_bayes_2023}.
We also show results for ridge and lasso estimators with a fixed $\tau=1$. Fixed ridge seems to perform the best amongst these models, and we use this model for our comparisons in Section \ref{rf}.

\begin{table*}[t]
      \centering
      \begin{tabular}{cccccccccc}
      \hline
      & \multicolumn{4}{c}{HLS vs Add-1} & & \multicolumn{4}{c}{HLS vs HDP} \\
        \multirow{2}{*}{BNC} & \multicolumn{2}{c}{Non-Bayesian} & \multicolumn{2}{c}{Bayesian} & & \multicolumn{2}{c}{Non-Bayesian} & \multicolumn{2}{c}{Bayesian}\\
      & 0/1 & Log & 0/1 & Log & & 0/1 & Log & 0/1 & Log \\ \hline
      TAN    & 34-4-12  & 33-0-17  & 30-4-16  & 30-0-20  & & 25-4-13  & 31-0-11  & 24-2-16  & 31-0-11 \\
      kDB-1  & 30-3-17  & 33-0-17  & 24-3-23  & 32-0-18  & & 24-3-15  & 31-0-11  & 22-2-18  & 31-0-11 \\
      kDB-2  & 27-4-19  & 34-0-16  & 24-3-23  & 37-0-13  & & 20-3-19  & 25-0-17  & 21-1-20  & 26-0-16 \\
      kDB-3  & 29-4-17  & 36-0-14  & 28-4-18  & 41-0-9   & & 24-5-13  & 27-0-15  & 27-1-14  & 27-0-15 \\ \hline
      \end{tabular}
    \caption{Win-Draw-Loss records for HLS versus Add-1 and HDP, on TAN and kDB with $k\in\{1,2,3\}$, under zero-one (0/1) loss and log loss.}
    \label{fig:bncs}
\end{table*}

\begin{figure*}[!b]
     \centering
     \begin{subfigure}[b]{0.4\textwidth}
         \centering
         \includegraphics[width=\textwidth]{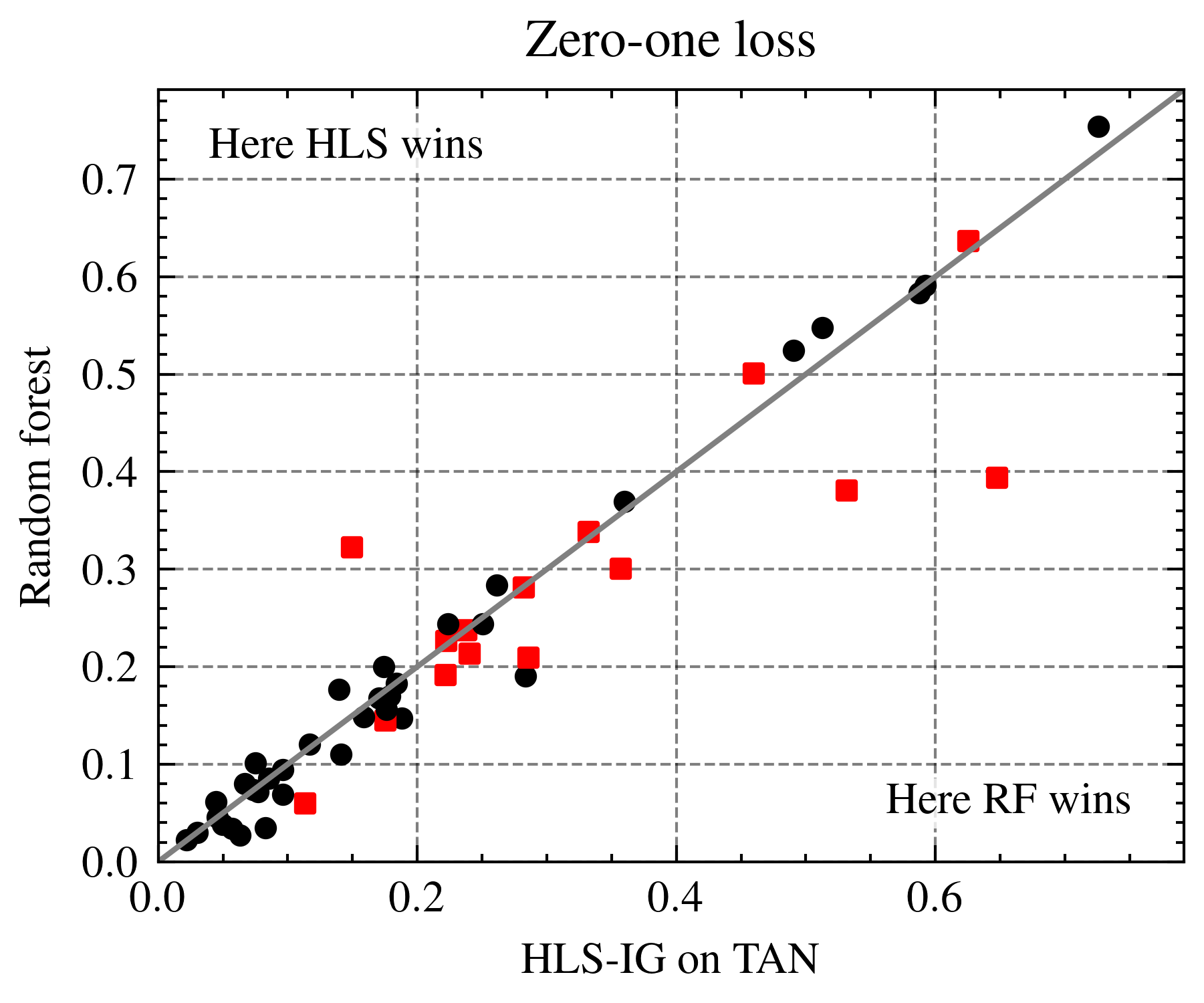}
         \caption{}
         \label{fig:rf1}
     \end{subfigure}
     \begin{subfigure}[b]{0.4\textwidth}
         \centering
         \includegraphics[width=\textwidth]{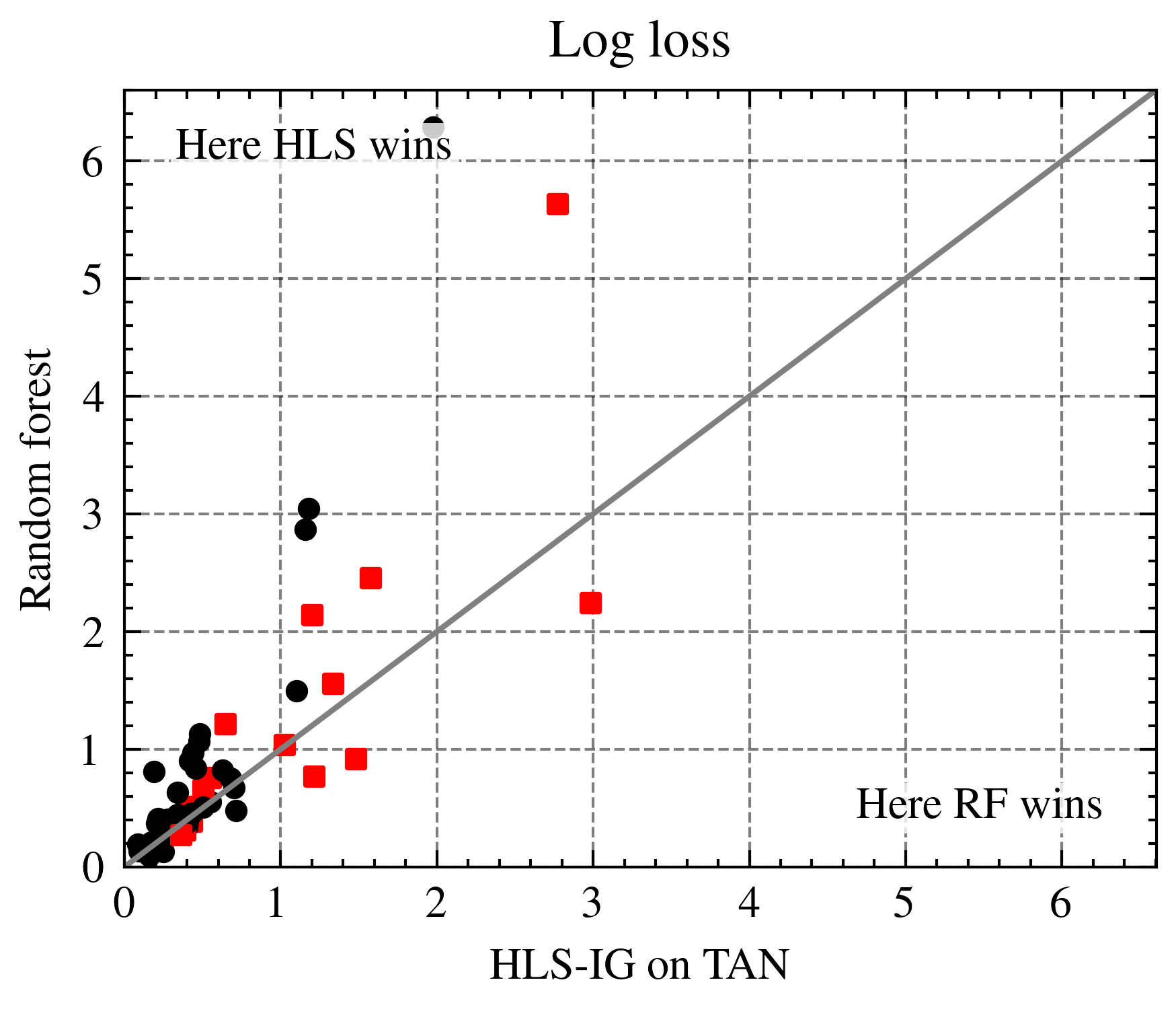}
         \caption{}
         \label{fig:rf2}
     \end{subfigure}
     \caption{Scatter plots for HLS-IG on TAN vs random forests, under (a) zero-one loss and (b) log loss. Red squares indicate datasets with top-15 cross-validation variance.}
     \label{fig:rf}
\end{figure*}

\subsubsection{Bayesian} \label{bayes}
Despite cross-validation failing to provide a well-tuned regularization parameter, it might be possible to extract better performance through a fully Bayesian procedure.
This section uses ridge
estimators under the global-local shrinkage hierarchy (see Section \ref{glsh}), where there are no local shrinkage parameters.

We first use a standard half-Cauchy prior $C^+(0, 1)$ for the global shrinkage parameters, as is common in the literature (see, for example, \citealp{polson_half-cauchy_2011}, \citealp{makalic_high-dimensional_2016}, \citealp{carvalho_horseshoe_2010}).
This method, perhaps surprisingly, does not perform as well as a simple fixed $\tau=1$ (see Table \ref{fig:regularization}).

However, as \citet{polson_half-cauchy_2011} discuss, it is not necessarily true that something as heavy-tailed as the half-Cauchy is appropriate for logistic regression (as opposed to regular linear regression).
Therefore we experiment with the inverse-gamma prior as an alternative. We show results for $\text{IG}(1/2, 1/2)$.
In Table \ref{fig:regularization}, we find that this setup outperforms non-Bayesian ridge regression, but in Figure \ref{fig:igvnb1} we see that the difference in performance is relatively minor.
Results for horseshoe models are available in Appendix A.4, but do not show any improvement.

To summarise: our best performing model uses Bayesian ridge regression under the GLS hierarchy, with inverse-gamma priors on the global shrinkage parameter and a P\'olya-Gamma scale mixture to represent the logistic target.
Nevertheless, non-Bayesian logistic ridge regression still provides a strong baseline and is much more commonly available in programming libraries.
These experiments highlight the flexibility of HLS; there is plenty of room to leverage theory on linear models to potentially improve performance.

\begin{table*}[t]
  \centering
  \begin{tabular}{cccccc}
  \hline
  \multicolumn{6}{c}{Win-Draw-Loss for BNCs vs random forests} \\
  Structure & Loss & Add-1 & HDP   & HLS-NB & HLS-IG \\
  \multirow{2}{*}{TAN}
            & 0/1-Loss & 16-5-29 & 15-5-25 & 20-4-26 & 19-4-27 \\
            & Log Loss & 34-0-16 & 31-0-11 & 35-0-15 & 36-0-14 \\ \hline
  \multirow{2}{*}{kDB-3}
            & 0/1-Loss & 11-4-35 & 10-2-30 & 18-2-30 & 15-4-31 \\
            & Log Loss & 31-0-19 & 29-0-13 & 33-0-17 & 33-0-17 \\ \hline
  \end{tabular}
  \caption{Win-Draw-Loss records against random forests for BNCs with additive smoothing, HDP, and HLS.}
  \label{fig:rftable}
\end{table*}

\begin{figure*}[!b]
     \centering
     \begin{subfigure}[b]{\textwidth}
         \centering
         \includegraphics[width=0.9\textwidth]{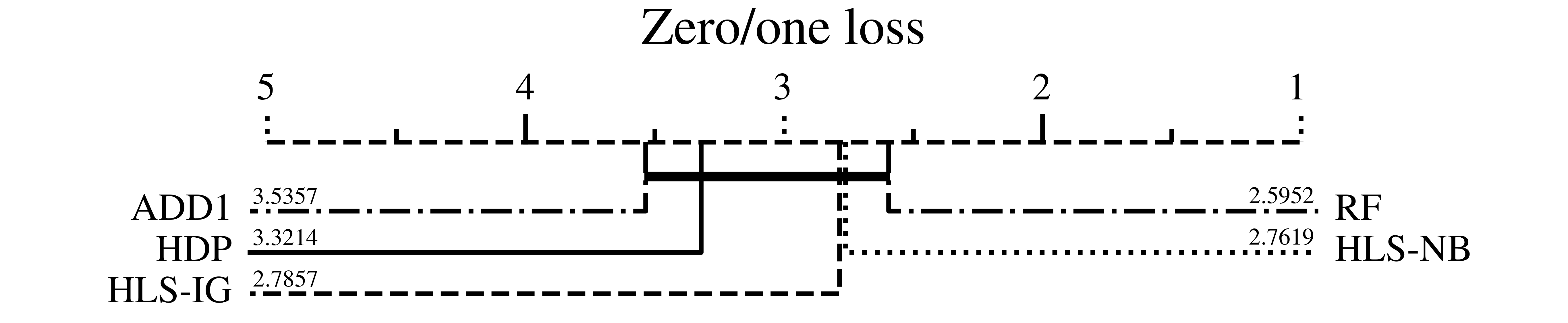}
     \end{subfigure}
     \begin{subfigure}[b]{\textwidth}
         \centering
         \includegraphics[width=0.9\textwidth]{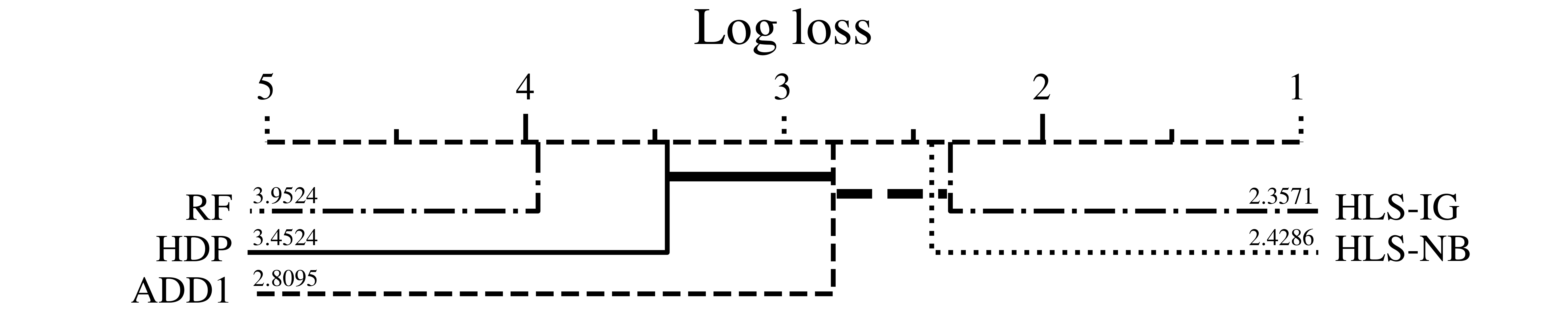}
     \end{subfigure}
     \caption{Critical difference diagrams representing the mean rank of models under zero-one loss and log loss, across the 42 datasets that HDP was trained on. BNCs use TAN as a structure.}
     \label{fig:cd}
\end{figure*}

\subsection{HLS performance} \label{performance}
In this section, we see how HLS compares to models for classification: additive smoothing and HDP smoothing for BNCs, and random forests (RF).
We do this using two variants of HLS: the standard non-Bayesian ridge in Scikit-learn (which we will call HLS-NB), and our best-performing Bayesian ridge (which we will call HLS-IG), using an inverse-gamma prior on the global shrinkage parameters.

\subsubsection{HLS vs BNCs on UCI} \label{bncs}
We compare BNCs with HLS-IG to BNCs with both HDP smoothing and additive smoothing.
The aim is to show
(a) how HLS compares to a simple baseline in the form of additive smoothing;
and (b) that HLS has similar performance to HDP smoothing, as intended.
For these experiments, we use the following structures: TAN, kDB-1, kDB-2, and kDB-3.
Further details of results are available in the appendices.

For additive smoothing, we present results in Table \ref{fig:bncs} for $m=1$ (Add-1) (where $m$ is the number of pseudo-counts; see Section \ref{parameterlearning}. This constant was chosen because it generally performed the best out of a variety of options.
We see that HLS-IG largely outperforms additive smoothing.
Results for further choices of $m$ are available in Appendix A.2.
It should also be noted that additive smoothing is known to significantly outperform maximum likelihood estimates, even for a relatively small $m$ (\cite{petitjean_accurate_2018}). 

For HDP smoothing, we only experiment on a subset (of size 42) of our collection of datasets, due to issues with the implementation.
Results in Figure \ref{fig:bncs} suggest that HLS can actually surpass the performance of HDP smoothing, although most wins are marginal (see Appendix A.3).

\subsubsection{BNCs vs random forests on UCI} \label{rf}
Now we compare the performance of BNCs with HLS to random forests.
In Table \ref{fig:rftable}, we present results using both TAN and kDB-3 as structures, comparing random forests to all of Add-1, HDP smoothing, HLS-NB and HLS-IG.
Interestingly, all BNCs outperform RF in terms of log loss, but lose in terms of zero-one loss. However, both HLS models offer improvements in zero-one loss compared to other BNCs, relative to random forests. 
In Figure \ref{fig:rf}, we see that there are significant wins on either side. These results show that BNCs with HLS can be competitive with random forests, and even preferable depending on what is desired.

In Figure \ref{fig:cd}, we present critical difference diagrams (\citealt{demsar_statistical_2006}; code from \citealt{IsmailFawaz2018deep}) for our models.
Under log loss, HLS ranks higher than HDP, and all BNCs rank higher than random forests with statistical significance.
Under zero-one loss, the difference between RF and the other models is not statistically significant.

\begin{table}[t]
  \centering
  \begin{tabular}{ccccc}
  \\\hline
  Cardinality & 2 & 2 & 5 & 5 \\
  Samples & 1000 & 100000 & 1000 & 100000 \\
  \hline
  HLS-fast & 0.0015 & 0.0284 & 0.0041 & 0.2637 \\
  HLS-slow & 0.0021 & 0.0353 & 0.3614 & 0.7495 \\
  HDP      & 0.2585 & 1.7554 & 1.8053 & 16.442 \\ \hline
  \end{tabular}
  \caption{Execution time (seconds) for methods on random categorical data. The time recorded is the minimum over 10 repetitions.}
  \label{fig:scaling}
\end{table}

\subsection{Timing experiments} \label{timing}
In this section we measure the time taken for our various methods on real and simulated data. All experiments were performed on a single core using an AMD Ryzen 5600X processor. 
We emphasise that these results are not rigorous:
Our HLS implementations are not fully optimized, and there is some additional overhead with HDP as it calls Java code from Python.

First, we compare the parameter learning time of HLS and HDP directly by training trees for nodes with 4 random parents, varying the cardinality and number of samples. Data is generated from a discretized multivariate normal distribution.
HLS-slow is the same Scikit-learn (Pedregosa et al., 2011) implementation we use in our experiments. As this implementation requires categorical counts (see section 3.2), we also record the time taken if you exclude constructing the design matrix with duplicate counts (HLS-fast).
Results are presented in Table \ref{fig:scaling}. We see that HLS is able to be orders of magnitude faster than HDP smoothing.

For real data, we use TAN as the structure and compare HLS-NB to HDP, Add-1 and RF. The results for the 42 datasets (10 folds each) that HDP was tested are as follows:
\begin{itemize}
\item Learning the structure (TAN) took c. 115 seconds
\item Add-1 took c. 79 seconds (excluding learning the structure, i.e. 115+79 seconds total)
\item HLS-NB took c. 87 seconds (excluding the structure)
\item HDP took c. 865 seconds (excluding the structure)
\item RF took c. 71 seconds
\end{itemize}
These results are essentially what one would expect; RF is the fastest overall and Add-1 is the fastest BNC, but HLS-NB only costs a small amount of time to achieve the performance of HDP. Note that the gap between HLS-NB/HDP and Add-1/RF will increase with larger BNC structures.

\section{Conclusion}
In this paper, we presented a novel method for learning smooth estimates of the
probabilities in Bayesian network classifiers (BNCs). It does this by training a
log-linear model over coefficients representing the $n$-th order interactions between predictors, for each $1\leq n \leq p$.
In contrast to other methods for jointly learning the parameters of BNC nodes, our formulation enables BNCs to leverage the simplicity and speed of linear models, with potential for more flexible and interpretable smoothing behaviour.
We demonstrated the effectiveness of HLS with TAN and kDB on a variety of categorical datasets from the UCI archive, compared with other BNCs and random forests.
We believe that this is a step towards making BNCs a viable contender to other state-of-the-art classification methods.
In future work, we wish to 
extend HLS to methods that scale to larger cardinalities, or support continuous features directly. 
We would also like to look at more discriminative design matrices, and extend HLS to a broader class of models that can benefit from similar hierarchical structures.


\pagebreak
\appendix

\clearpage

\section{Additional Experiments}\label{appendix:experiments}

\subsection{HLS intercept}

\begin{figure}[H]
     \centering
     \begin{subfigure}[b]{0.45\textwidth}
         \centering
         \includegraphics[width=\textwidth]{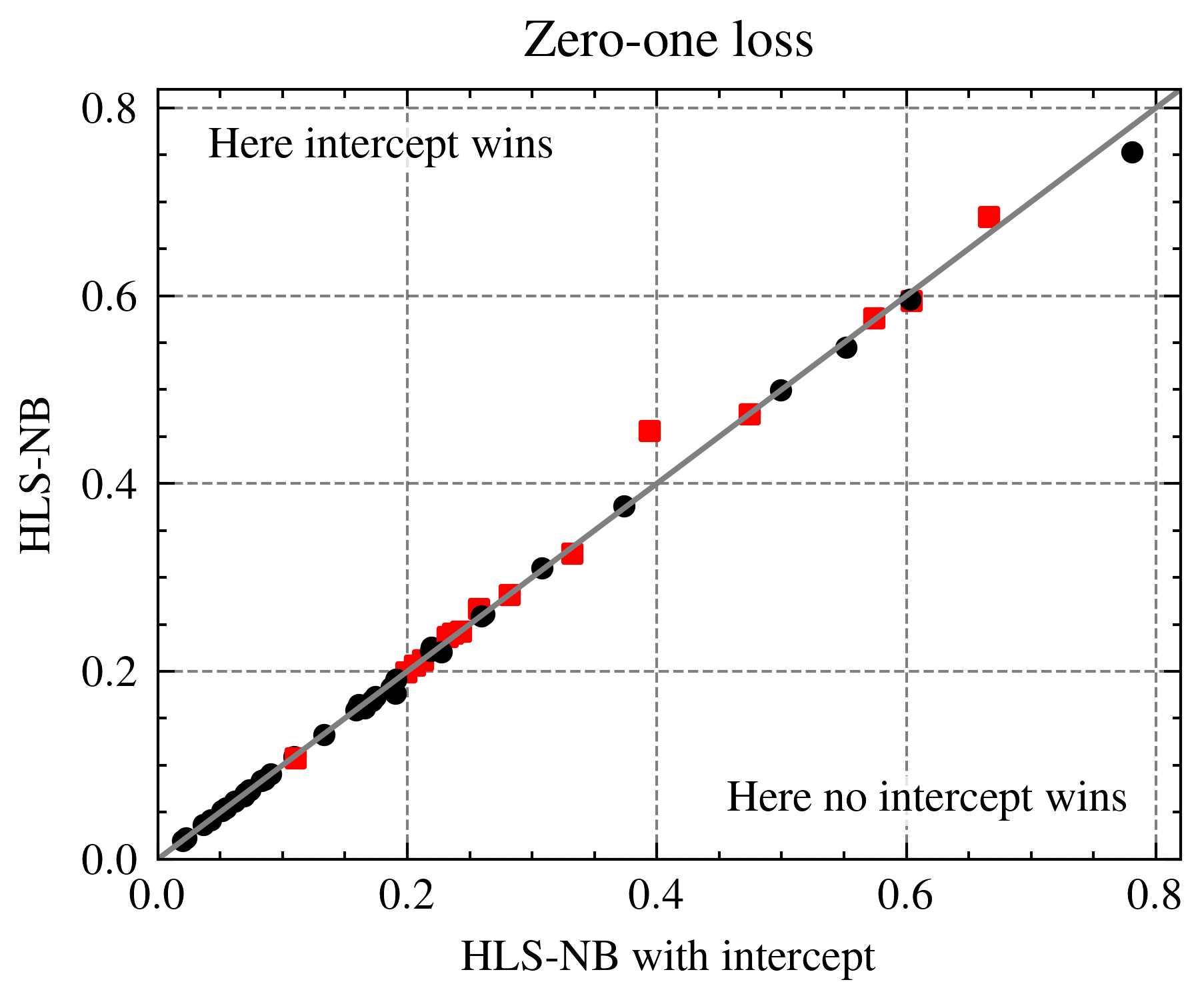}
         \caption{W-D-L: 18-13-19}
     \end{subfigure}
     \begin{subfigure}[b]{0.45\textwidth}
         \centering
         \includegraphics[width=\textwidth]{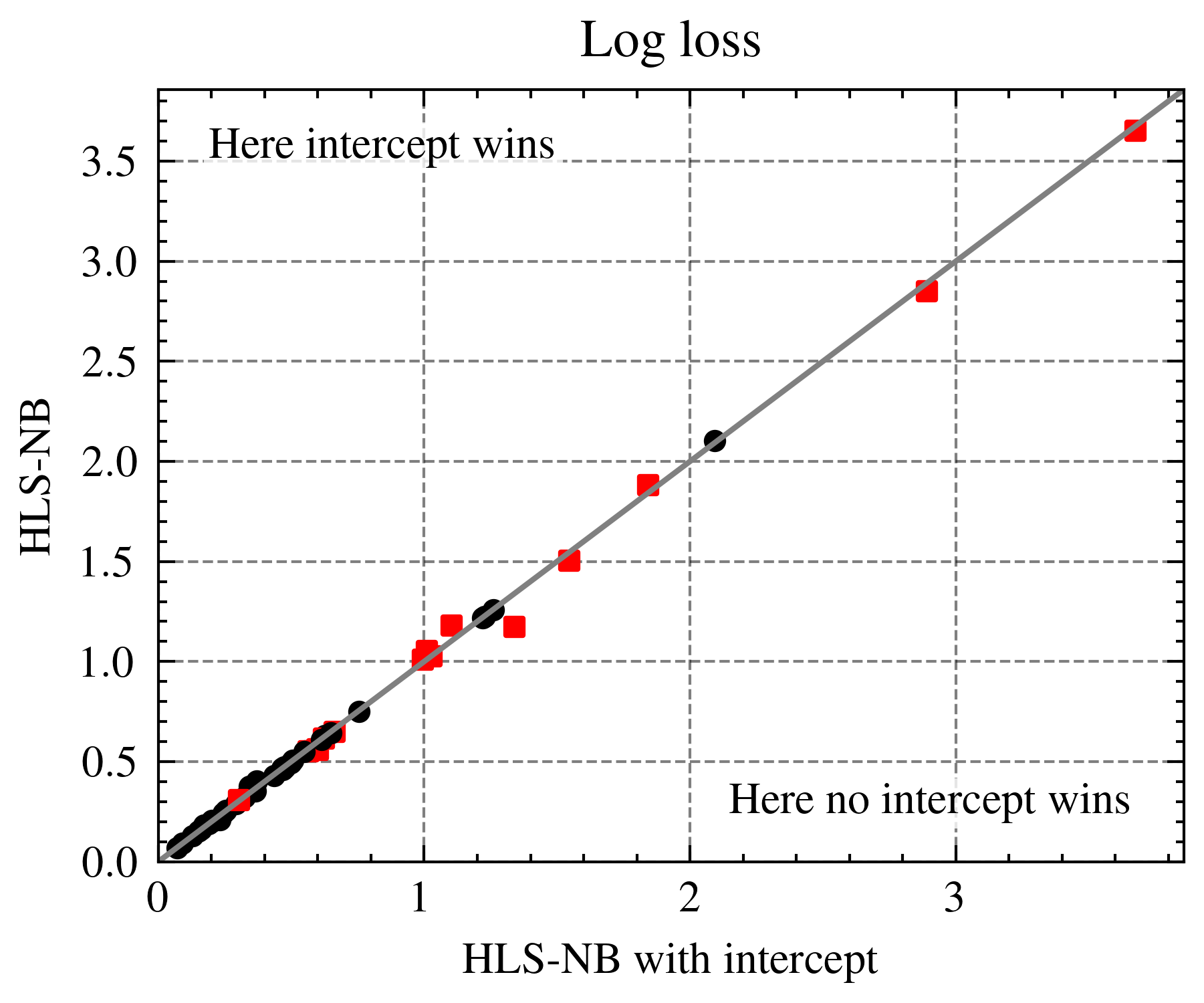}
         \caption{W-D-L: 20-0-30}
     \end{subfigure}
     \caption{Scatter plots for HLS-NB with an intercept vs HLS-NB with no intercept, on kDB-3, under (a) zero-one loss and (b) log loss. Red squares indicate datasets with top-15 cross-validation variance.}
     \label{fig:intercept}
\end{figure}

We present results in Figure \ref{fig:intercept} for HLS-NB with an intercept in the regression, versus HLS-NB without an intercept. 
On this data, we see virtually no difference between the two classifiers.
In the main sections of this paper, we choose to use no intercept due to its simplicity.

\subsection{Additive smoothing}
\begin{figure}[H]
     \centering
     \begin{subfigure}[b]{0.45\textwidth}
         \centering
         \includegraphics[width=\textwidth]{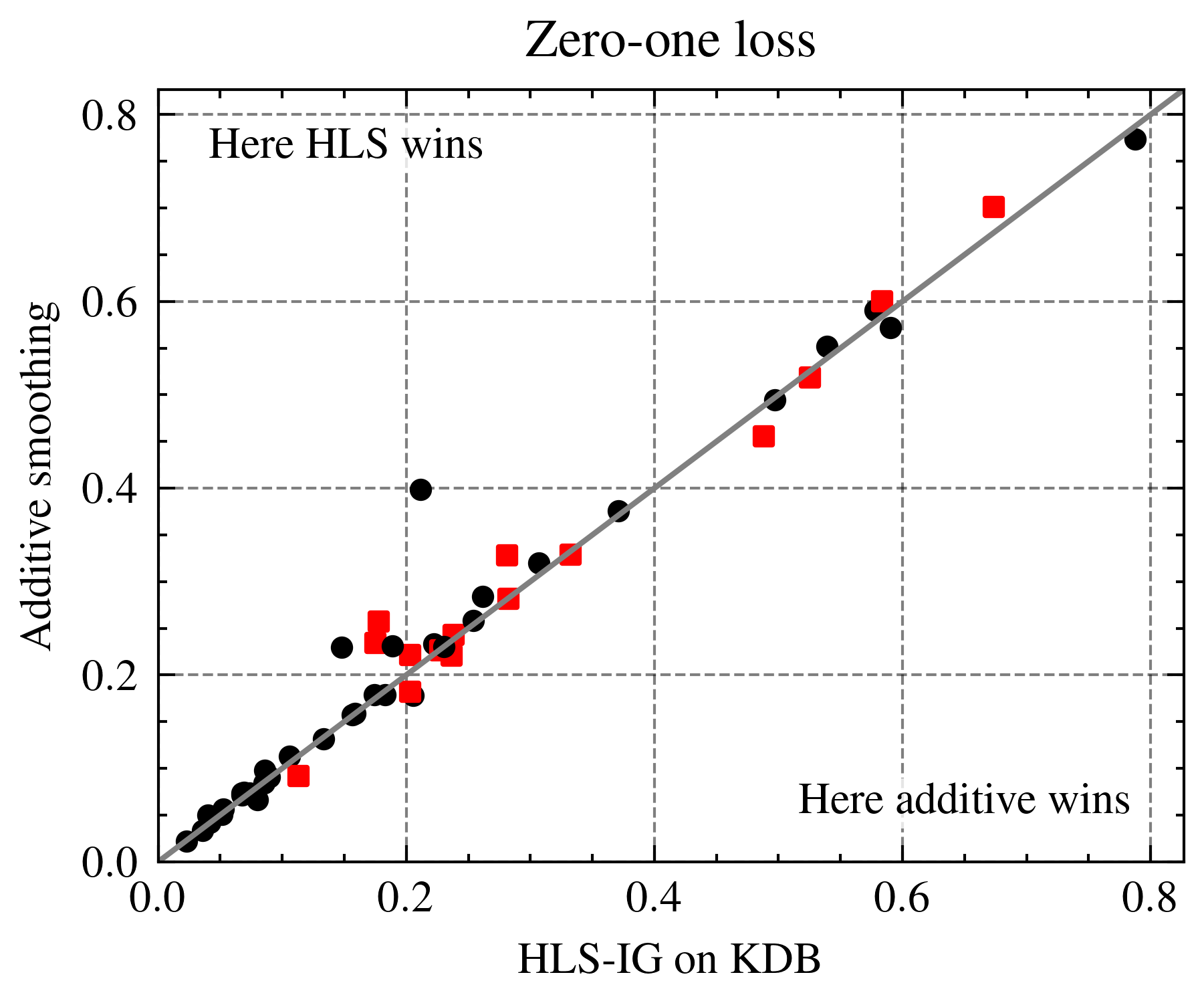}
         \caption{W-D-L: 28-4-18}
     \end{subfigure}
     \begin{subfigure}[b]{0.45\textwidth}
         \centering
         \includegraphics[width=\textwidth]{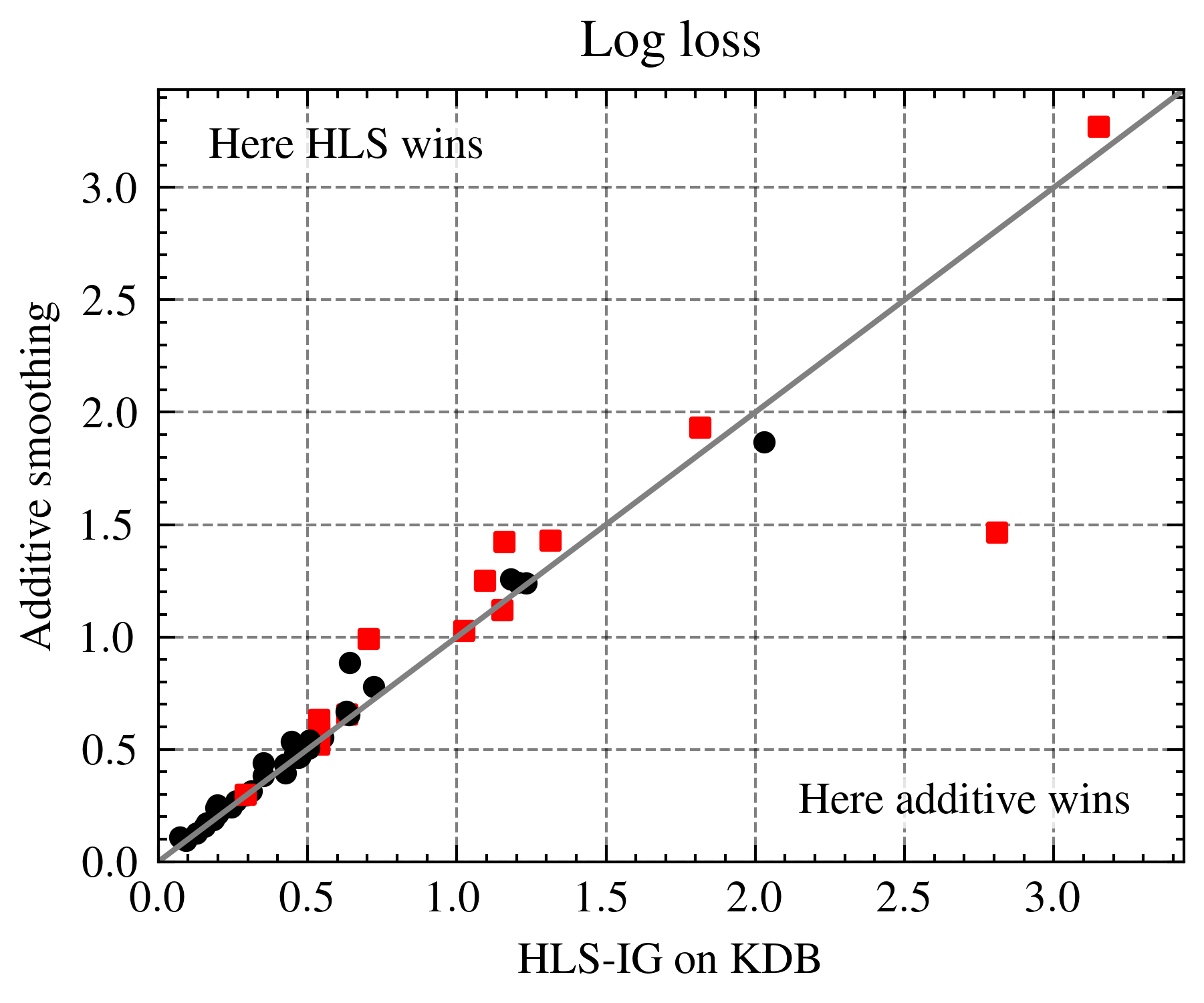}
         \caption{W-D-L: 41-0-9}
     \end{subfigure}
     \caption{Scatter plots for HLS-IG on TAN vs Add-1, under (a) zero-one loss and (b) log loss. Red squares indicate datasets with top-15 cross-validation variance.}
     \label{fig:add1}
\end{figure}

In Table \ref{fig:addtable}, we present results for various choices of pseudo-counts $m$ in additive smoothing (Add-$m$) on kDB-3, and
Figure \ref{fig:add1} shows scatter plots for Add-1 versus HLS-IG.
HLS achieves significant improvements over additive smoothing on many of the datasets, while losses are small (points below the diagonal in Figure \ref{fig:add1}).
As mentioned in Section \ref{nonbayes}, using cross-validation to select $m$ can significantly reduce performance on these datasets.

\begin{table}[H]
  \centering
  \begin{tabular}{ccccccc}
  \\\hline
  \multicolumn{7}{c}{Win-Draw-Loss for Add-$m$ vs HLS-NB} \\
  Loss & $m=0.2$ & $m=0.5$ & $m=1$ & $m=2$ & $m=5$ & $m=20$ \\
  0/1-Loss & 10-4-36 & 10-5-35 & 17-4-29 & 23-2-25 & 21-3-26 & 18-3-29 \\
  Log Loss & 6-0-44 & 7-0-43  & 14-0-36 & 20-0-30 & 21-0-29 & 19-0-31 \\ \hline
  \end{tabular}
  \caption{Win-Draw-Loss records for additive smoothing against HLS-NB}
  \label{fig:addtable}
\end{table}


\subsection{HLS vs HDP}
\begin{figure}[H]
     \centering
     \begin{subfigure}[b]{0.45\textwidth}
         \centering
         \includegraphics[width=\textwidth]{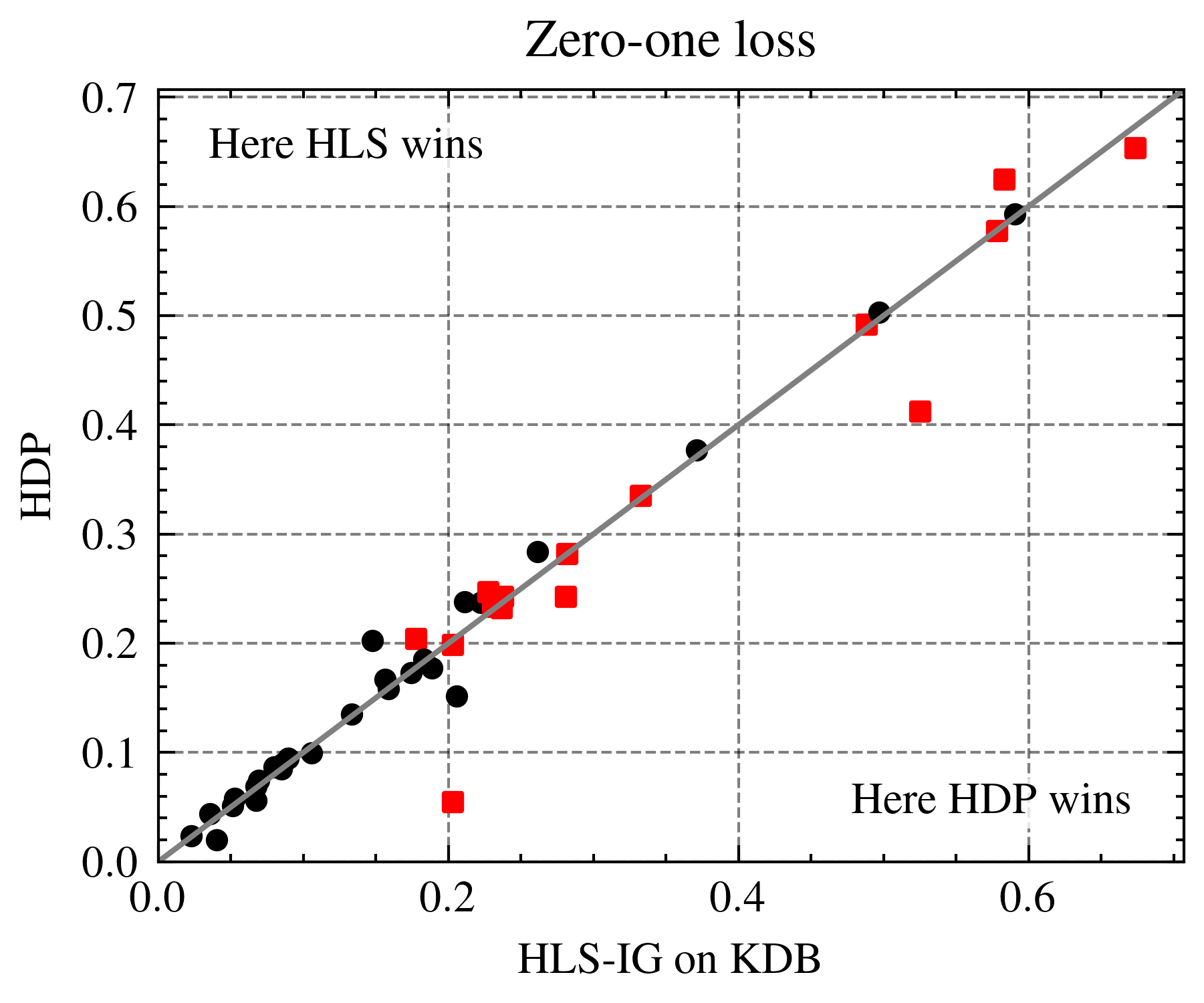}
         \caption{W-D-L: 27-1-14}
     \end{subfigure}
     \begin{subfigure}[b]{0.45\textwidth}
         \centering
         \includegraphics[width=\textwidth]{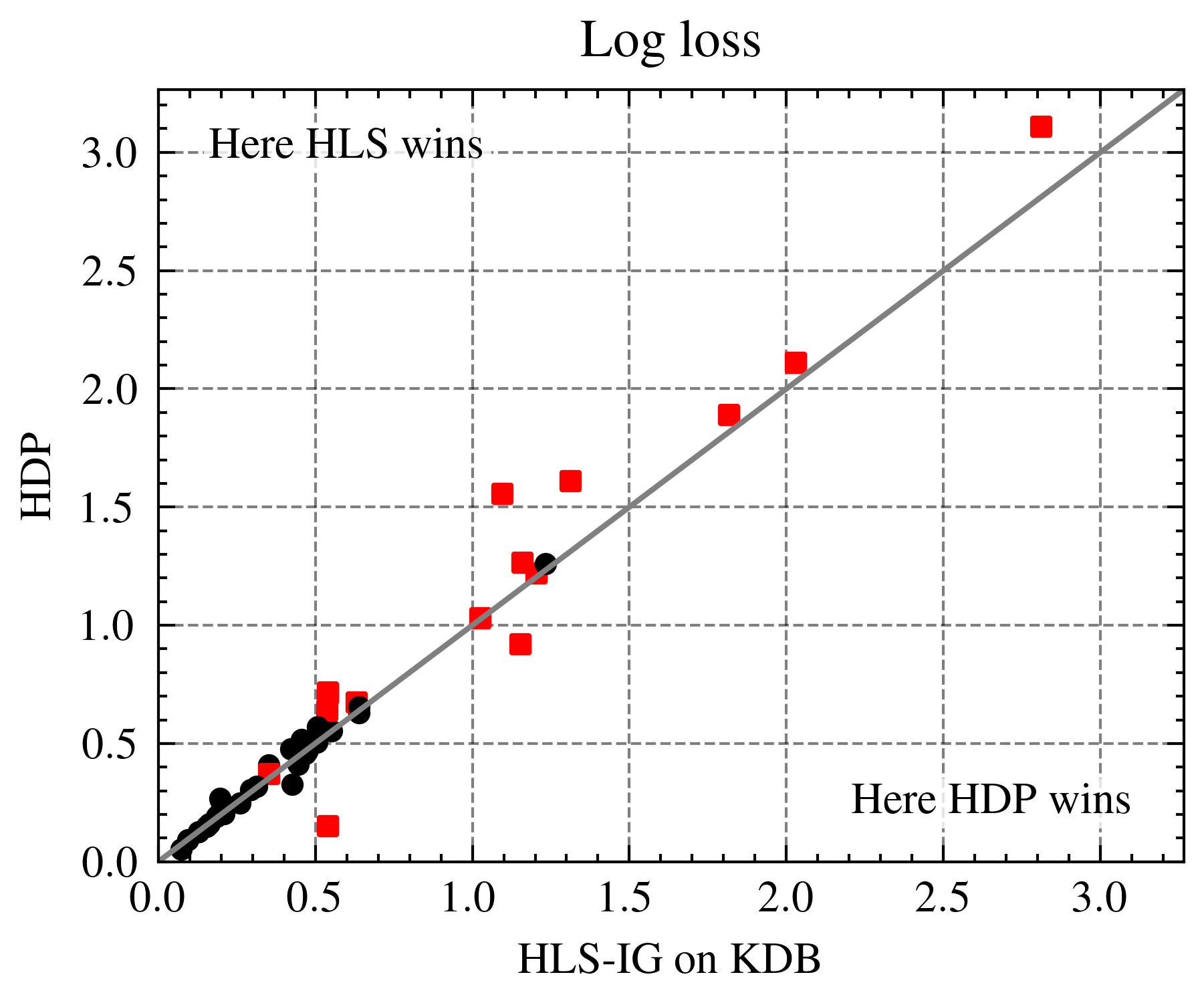}
         \caption{W-D-L: 27-0-15}
     \end{subfigure}
     \caption{Scatter plots for HLS-IG on TAN vs HDP on 42 datasets, under (a) zero-one loss and (b) log loss. Red squares indicate datasets with top-15 cross-validation variance.}
     \label{fig:appendix-hdp}
\end{figure}

In Figure \ref{fig:appendix-hdp}, we present scatter plots for HDP smoothing versus HLS-IG (the same experiment as in Table \ref{bncs}). While HLS seems to win overall, there are some significant wins on either side on individual datasets.

\subsection{Horseshoe}

We experiment with two additional models under the GLS hierarchy:
\begin{itemize}
  \item The horseshoe (HS), where global shrinkage parameter $\tau\sim C^+(0, 1)$ and local shrinkage parameters $\lambda_j\sim C^+(0, 1)$;
  \item The horseshoe with inverse-gamma priors (HS-IG) on the shrinkage parameters, where $\tau\sim \text{IG}(1/2, 1/2)$ and $\lambda_j\sim \text{IG}(1/2, 1/2)$.
\end{itemize}

\begin{figure}[H]
     \centering
     \begin{subfigure}[b]{0.45\textwidth}
         \centering
         \includegraphics[width=\textwidth]{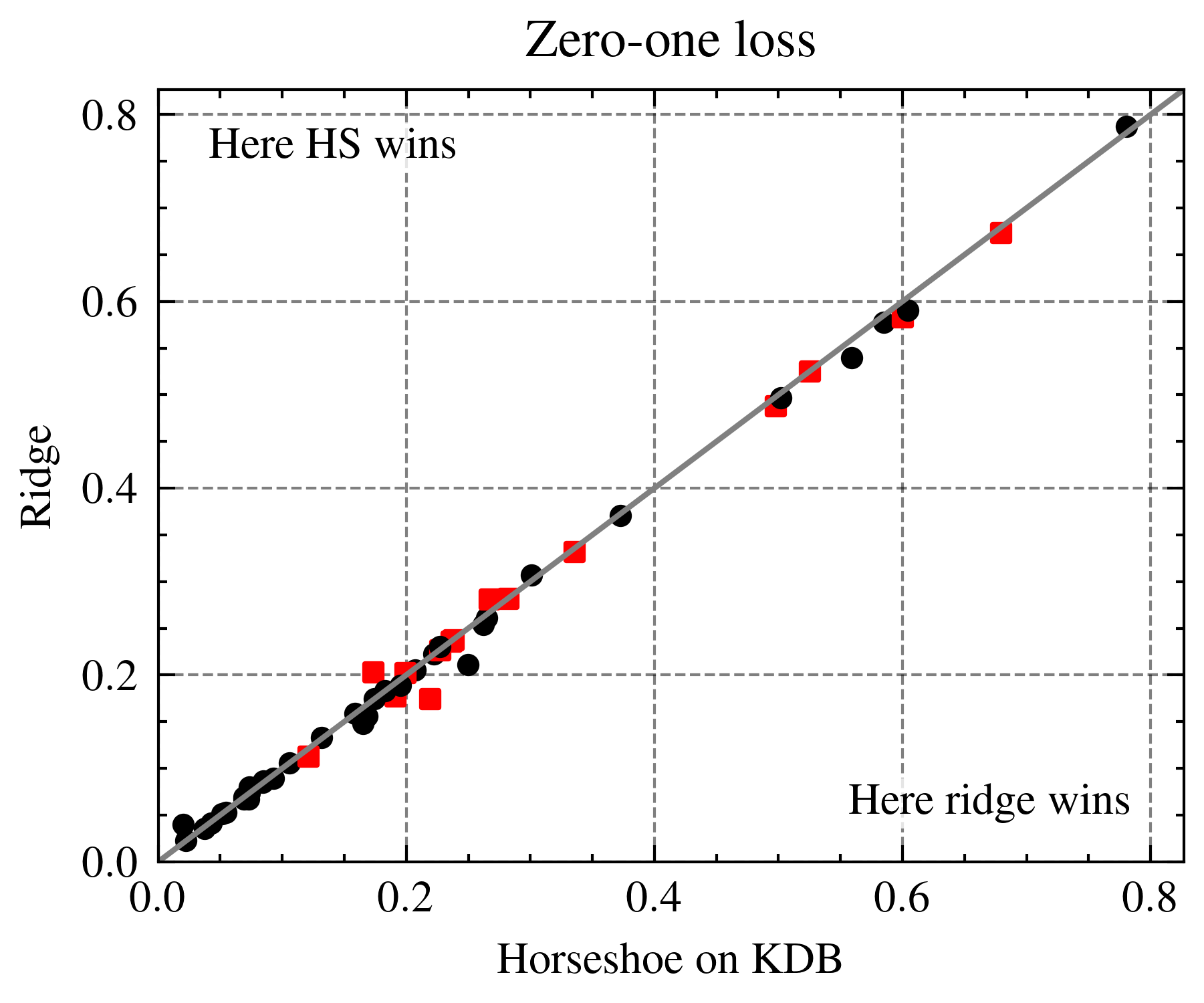}
         \caption{}
     \end{subfigure}
     \begin{subfigure}[b]{0.45\textwidth}
         \centering
         \includegraphics[width=\textwidth]{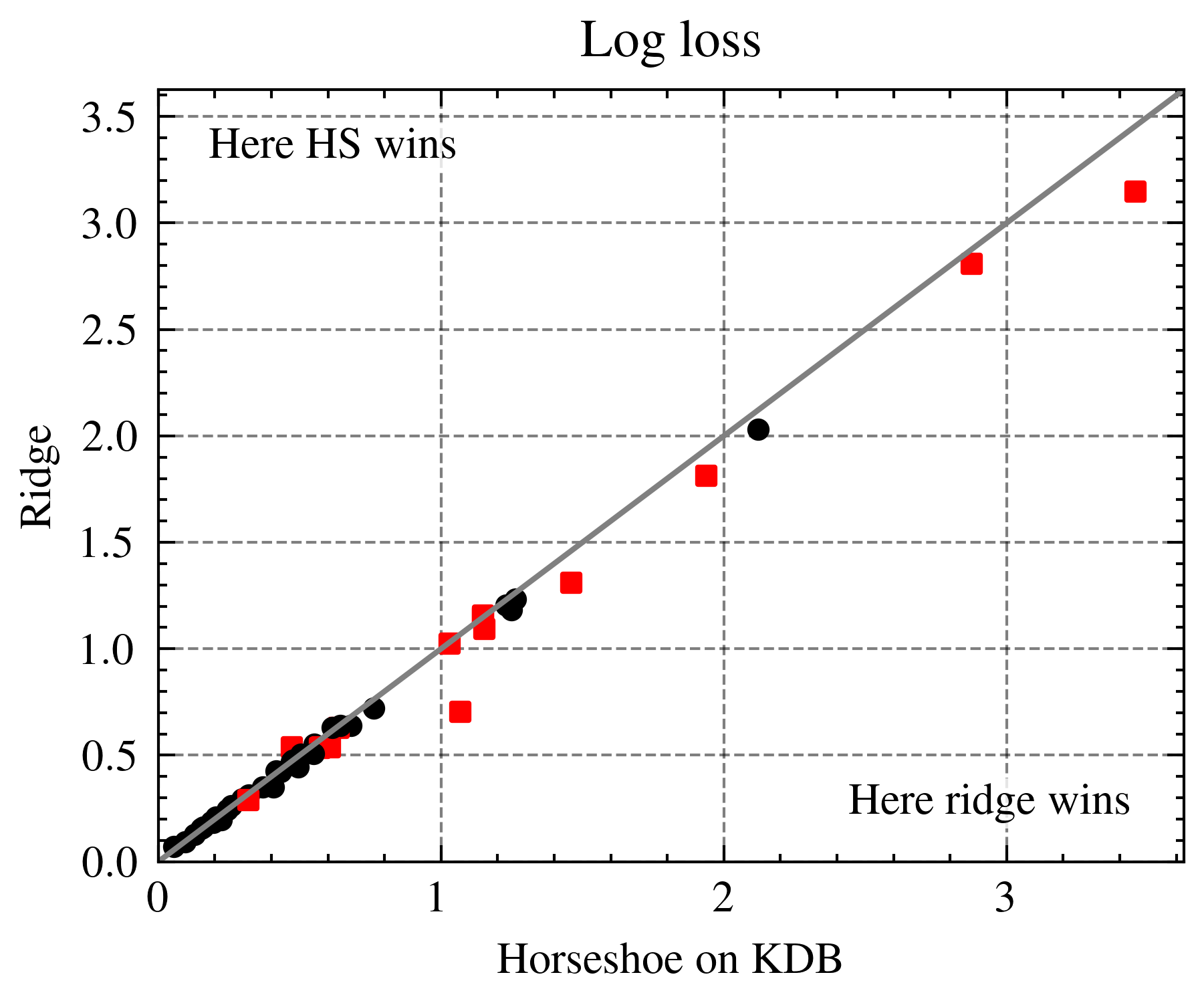}
         \caption{}
     \end{subfigure}
     \caption{Scatter plots for HLS with HS vs HLS-IG on kDB-3, under (a) zero-one loss and (b) log loss. Red squares indicate datasets with top-15 cross-validation variance.}
     \label{fig:hsc}
\end{figure}

\begin{figure}[H]
     \centering
     \begin{subfigure}[b]{0.45\textwidth}
         \centering
         \includegraphics[width=\textwidth]{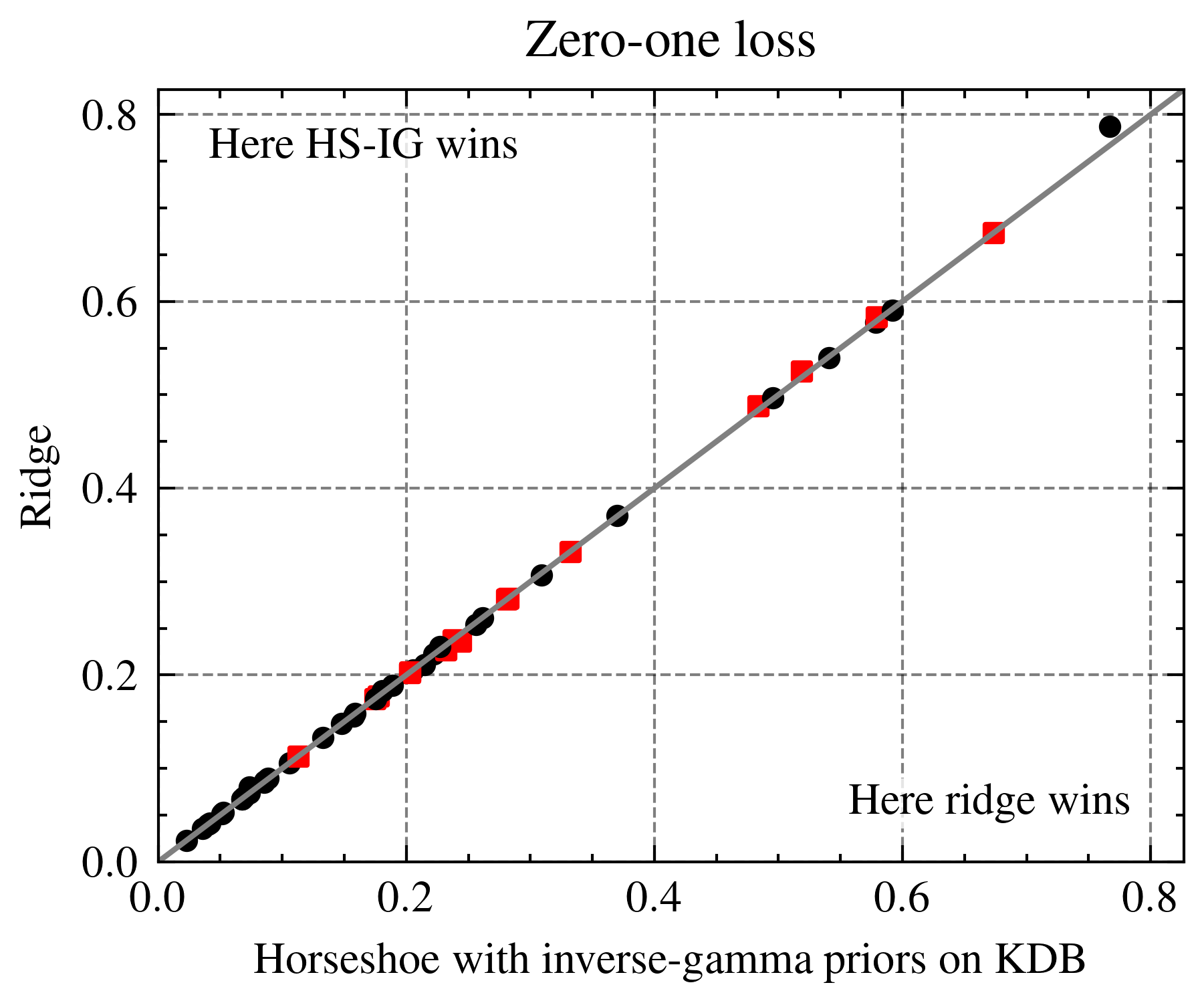}
         \caption{}
     \end{subfigure}
     \begin{subfigure}[b]{0.45\textwidth}
         \centering
         \includegraphics[width=\textwidth]{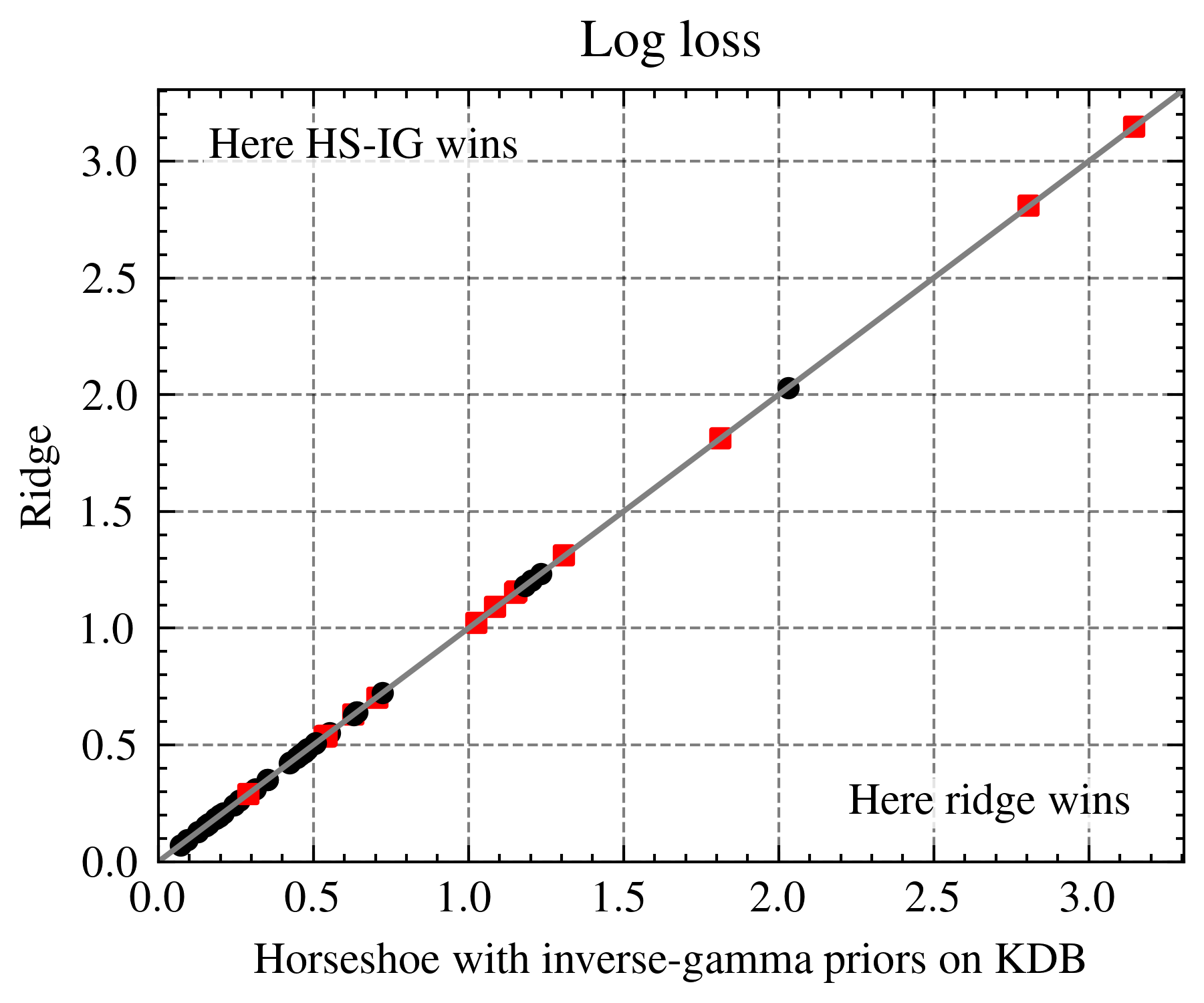}
         \caption{}
     \end{subfigure}
     \caption{Scatter plots for HLS with HS-IG vs HLS-IG on kDB-3, under (a) zero-one loss and (b) log loss. Red squares indicate datasets with top-15 cross-validation variance.}
     \label{fig:hsig}
\end{figure}

Results for kDB-3 are presented in Table \ref{fig:table3}.
Scatter plots for HS versus HLS-IG are presented in Figure \ref{fig:hsc} and HS-IG versus HLS-IG in Figure \ref{fig:hsig}.
Similar to the experiments in Section \ref{bayes}, we find that inverse-gamma priors perform better than Cauchy. In the scatterplot in Figure \ref{fig:hsig}, the inverse-gamma horseshoe performs very similarly to the ridge HLS-IG, but we prefer the latter due to its simplicity ($\lambda_j=1$).

\begin{table}[H]
  \centering
  \begin{tabular}{ccc}
  \\
  \multicolumn{3}{c}{Win-Draw-Loss for Add-$m$ vs HLS-NB} \\
  \hline
  Regularization & 0/1-Loss & Log loss \\
  Horseshoe with Cauchy & 15-8-27 & 13-0-37 \\
  Horseshoe with inverse-gamma & 15-18-17 & 26-0-24 \\ \hline
  \end{tabular}
  \caption{Win-Draw-Loss records for horseshoe models against HLS-IG}
  \label{fig:table3}
\end{table}

\clearpage

\section{P\'olya-Gamma Sampler} \label{appendix:sampler}

From Polson et al. (2013),
a random variable $X$ has a P\'olya-Gamma distribution with parameters $b>0$ and $c\in\mathbb{R}$, denoted $X\sim \text{PG}(b,c)$, if it is equal in distribution to
\[
  \frac{1}{2\pi^2} \sum_{k=1}^\infty\frac{g_k}{(k-1/2)^2 + c^2/(4\pi^2)},
\] where $g_k\sim \text{Gamma}(b, 1)$ are independent random variables.

To sample $X\sim \text{PG}(b, c)$,
we first sample the first $K$ terms of this series, and then approximate the remaining sum with an additional random variable $G$:
\[
  \frac{1}{2\pi^2}\sum_{k=1}^K\frac{g_k}{(k-1/2)^2 + c^2/(4\pi^2)} + G.
\]
$G$ is a gamma-distributed random variable sampled using mean $\mu_G$ and variance $\text{var}_G$ chosen such that the resulting $X$ has mean and variance matching a $\text{PG}(b,c)$ random variable.
With some numerical stability for extreme values of $c$, this results in
\[
  \mu_G = \begin{cases}
    b/4 &\text{if } c \leq 1e^{-3} \\
    b/(2c) &\text{if } c \geq 300 \\
    \frac{\exp(c)-1}{\exp(c)+1} \frac{b}{2c} &\text{otherwise}
  \end{cases},
\]
\[
  \text{var}_G \begin{cases}
    b/24  &\text{if } c \leq 1e^{-3} \\
    b/(2c^3) &\text{if } c \geq 300 \\
    \frac{b(\exp(2c) - 2c\exp(c) - 1)}{2c^3(\exp(c)+1)^2} &\text{otherwise}
  \end{cases}.
\]
In our experiments, we use $K=2$.

\section{Datasets}\label{appendix:datasets}

In Table \ref{fig:datasets}, we list the datasets used in our experiments. UCI ID indicates the dataset's ID in the UCI archive \citep{lichman_uci_2013}.

\begin{longtable}{rlrr}
  \toprule
    \textbf{ID} & \textbf{Name} & \textbf{Samples} & \textbf{Features} \\
  \midrule
  \endfirsthead

  \multicolumn{4}{l}{\small\slshape Continued from previous page}\\
  \toprule
    \textbf{ID} & \textbf{Name} & \textbf{Samples} & \textbf{Features} \\
  \midrule
  \endhead

  \midrule
  \multicolumn{4}{r}{\small\slshape Continued on next page}\\
  \endfoot

  \bottomrule
  \\
    \caption{List of datasets. * indicates datasets that HDP was not trained on.}
    \label{fig:datasets}
  \endlastfoot

12  & Balance Scale                                     & 625    & 4  \\
17  & Breast Cancer Wisconsin (Diagnostic)              & 569    & 30 \\
19  & Car Evaluation                                    & 1728   & 6  \\
30  & Contraceptive Method Choice                       & 1473   & 9  \\
42  & Glass Identification                              & 214    & 9  \\
43  & Haberman's Survival                               & 306    & 3  \\
44  & Hayes-Roth                                        & 160    & 4  \\
50  & Image Segmentation                                & 2310   & 19 \\
53  & Iris                                              & 150    & 4  \\
63  & Lymphography                                      & 148    & 19 \\
*69  & Molecular Biology (Splice-junction Gene Sequences)& 3190   & 60 \\
70  & MONK's Problems                                   & 432    & 6  \\
76  & Nursery                                           & 12960  & 8  \\
*78  & Page Blocks Classification                        & 5473   & 10 \\
94  & Spambase                                          & 4601   & 57 \\
95  & SPECT Heart                                       & 267    & 22 \\
96  & SPECTF Heart                                      & 267    & 44 \\
101 & Tic-Tac-Toe Endgame                               & 958    & 9  \\
107 & Waveform Database Generator (Version 1)           & 5000   & 21 \\
109 & Wine                                              & 178    & 13 \\
111 & Zoo                                               & 101    & 16 \\
145 & Statlog (Heart)                                   & 270    & 13 \\
174 & Parkinsons                                        & 197    & 22 \\
176 & Blood Transfusion Service Center                  & 748    & 4  \\
186 & Wine Quality                                      & 4898   & 11 \\
212 & Vertebral Column                                  & 310    & 6  \\
257 & User Knowledge Modeling                           & 403    & 5  \\
267 & Banknote Authentication                           & 1372   & 4  \\
277 & Thoracic Surgery Data                             & 470    & 16 \\
292 & Wholesale customers                               & 440    & 7  \\
327 & Phishing Websites                                 & 11055  & 30 \\
329 & Diabetic Retinopathy Debrecen                     & 1151   & 19 \\
372 & HTRU2                                             & 17898  & 8  \\
379 & Website Phishing                                  & 1353   & 9  \\
445 & Absenteeism at work                               & 740    & 19 \\
519 & Heart failure clinical records                    & 299    & 12 \\
529 & Early Stage Diabetes Risk Prediction              & 520    & 16 \\
545 & Rice (Cammeo and Osmancik)                        & 3810   & 7  \\
563 & Iranian Churn                                     & 3150   & 13 \\
572 & Taiwanese Bankruptcy Prediction                   & 6819   & 95 \\
582 & Student Performance on an Entrance Examination    & 666    & 11 \\
*697 & Predict students' dropout and academic success    & 4424   & 36 \\
759 & Glioma Grading Clinical and Mutation Features     & 839    & 23 \\
*827 & Sepsis Survival Minimal Clinical Records          & 110341 & 3  \\
850 & Raisin                                            & 900    & 7  \\
*856 & Higher Education Students Performance Evaluation  & 145    & 31 \\
*863 & Maternal Health Risk                              & 1013   & 6  \\
890 & AIDS Clinical Trials Group Study 175              & 2139   & 23 \\
*915 & Differentiated Thyroid Cancer Recurrence          & 383    & 16 \\
*936 & National Poll on Healthy Aging (NPHA)             & 714    & 14 \\
\end{longtable}

\bibliography{HLS}

\end{document}